\documentclass[letterpaper, 10 pt, conference]{ieeeconf}  

\IEEEoverridecommandlockouts                              

\usepackage{cite}
\usepackage{amsmath,amssymb,amsfonts}
\usepackage[linesnumbered,ruled,vlined]{algorithm2e}
\usepackage{graphicx}
\usepackage{mathrsfs}
\usepackage{textcomp}
\usepackage{xcolor}
\usepackage{pifont}
\usepackage{caption}
\usepackage[colorlinks,citecolor=blue]{hyperref}
\usepackage{multirow}
\title{\LARGE \bf
Efficient Graduated Non-Convexity for Pose Graph Optimization
}

\author{Wonseok Kang, Jaehyun Kim, Jiseong Chung, Seungwon Choi, and Tae-wan Kim*
\thanks{* Corresponding Author}
\thanks{This research is supported by the ‘Development of Autonomous Ship Technology (20200615)’ funded by the Ministry of Oceans and Fisheries, Korea.}
\thanks{The authors are with the Department of Naval Architecture and Ocean Engineering, Seoul National University (SNU), Seoul 08826, Republic of Korea. \texttt{\{eoid361, jaedalong, dntksdmfwmf, csw3575, taewan\}@snu.ac.kr}
}
}

\begin{document}

\maketitle
\thispagestyle{empty}
\pagestyle{empty}

\begin{abstract}

We propose a novel approach to Graduated Non-Convexity (GNC) and demonstrate its efficacy through its application in robust pose graph optimization, a key component in SLAM backends. Traditional GNC methods often rely on heuristic methods for GNC schedule, updating control parameter $\mu$ for escalating the non-convexity. In contrast, our approach leverages the properties of convex functions and convex optimization to identify the boundary points beyond which convexity is no longer guaranteed, thereby eliminating redundant optimization steps in existing methodologies and enhancing both speed and robustness. We show that our method outperforms the state-of-the-art method in terms of speed and accuracy when used for robust back-end pose graph optimization via GNC. Our work builds upon and enhances the open-source riSAM framework. Our implementation can be accessed from: \href{https://github.com/SNU-DLLAB/EGNC-PGO}{https://github.com/SNU-DLLAB/EGNC-PGO}

\end{abstract}

\section{Introduction}

Simultaneous Localization and Mapping (SLAM) is a key area of research essential for the navigation of autonomous robots. The SLAM process is generally divided into two parts: the front-end, which processes sensor data to create initial movement estimates, and the back-end, which refines these estimates. Most SLAM approaches utilize pose graph optimization (PGO) techniques on the back-end. However, PGO is sensitive to front-end errors and unclear data, making effective outlier management crucial for reliable SLAM performance.

The urgency to improve SLAM performance is underscored by its growing applications in real-world scenarios. In autonomous vehicles, for instance, even minor errors in SLAM can result in significant safety risks. Similarly, in industrial robotics, SLAM accuracy can be directly tied to operational efficiency. Thus, there is a pressing need for more reliable and efficient SLAM systems.

Among these efforts, one such method is the Graduated Non-Convexity (GNC) based approach, which incrementally increases the non-convexity of an adjustable kernel function for more robust SLAM optimization. The GNC method enhances robustness by starting with a simplified convex problem and iteratively introducing the more complex, non-convex aspects of the original problem. This transition allows for better convergence and outlier management, with lower dependency on the initial estimation. While these GNC-based methods outperform existing techniques, they often rely on heuristic approaches and empirically derived strategies for increasing non-convexity, leaving room for further refinement.

In this paper, we propose a novel \textit{GNC schedule}, which is a strategy for elevating non-convexity in GNC-based optimization. By exploiting the convex properties of the kernel function, we optimize within the convex stable regions to reduce the number of iterations required. As a result, we achieve enhanced computational efficiency without sacrificing accuracy, thereby improving the performance of the base riSAM. Additionally, we plan to release the open-source code, built upon the existing riSAM \cite{mcgann2023robust} framework, to encourage further research in this domain.

The remainder of this paper is organized as follows: Section II reviews related work in the field, Section III introduces our methodology, Section IV presents the experimental methods and results that demonstrate the effectiveness of our approach, and Section V offers a discussion and concluding remarks.

\begin{figure}[!t]
    \includegraphics[width=0.486\textwidth]{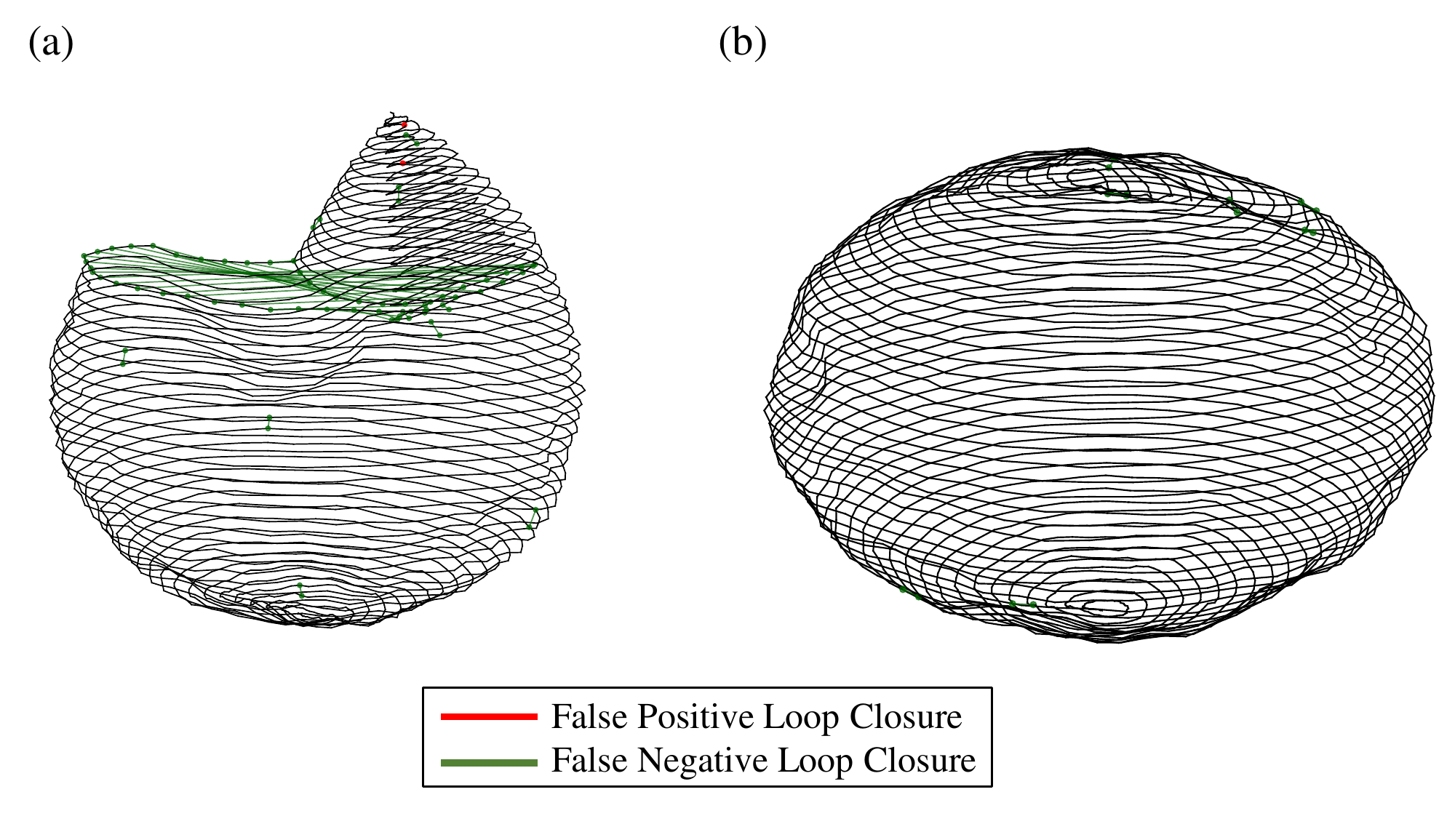}
    \caption{Results of an experimental study using the Sphere2500 dataset with 50\% ratio of false loop closure. (a) The result from riSAM shows that it fails to filter out some of the false positive loops, leading to a trajectory crash. (b) In contrast, the result of our proposed method indicates that it effectively addresses this challenge and has improved accuracy.}
    \label{fig:sphere_results}
\end{figure}

\section{Related Work}

\begin{figure*}[t]
    \includegraphics[width=\textwidth]{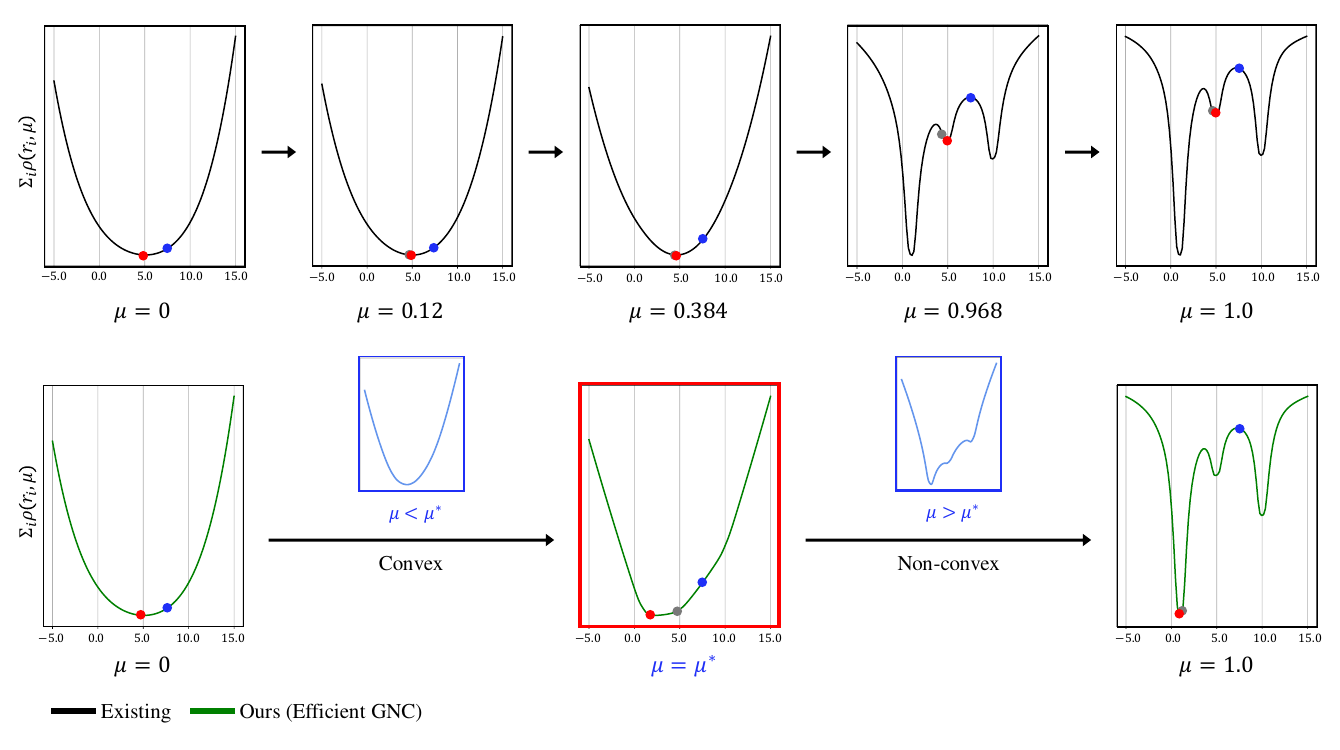}
    \caption{Comparison of the GNC optimization process between our proposed method and riSAM. For the simple linear regression example, the optimization process of riSAM is shown at the top, while our proposed method is shown at the bottom. The linear regression is conducted based on inlier data points (55\%) sampled from the original function $y=x$, outlier data points (15\%) sampled from $y=5x$, and another set of outlier data points (30\%) sampled from $y=10x$. In each graph, the initial guess (blue dot) is set to 7.5, and the previous updated value (gray dot), and the current updated value (red dot) are indicated. In the initial stages of GNC optimization, even as \cite{mcgann2023robust} adheres to the update rule given by $\mu_{i+1} = min (1.0, \mu_i + 1.2(\mu_i - \mu_{init} + 0.1))$, the convexity of the total cost function remains unchanged. Our proposed method preserves robustness with only one intermediate step(red box) for a specific $\mu$ value ($\mu^*$) inferred adaptively based on the residual of each data point. If the $\mu$ value is less than $\mu^*$(left blue box), the graph might become "too convex", leading to significant deviations from the original function. On the other hand, if the $\mu$ value exceeds $\mu^*$(right blue box), the graph could contain a non-convex part, risking convergence to local minima.}
    \label{fig:step_compare}
\end{figure*}

The effective handling of outliers is critical for achieving reliable solutions in PGO, a complex non-linear least squares problem. Over the years, a multitude of methodologies have been put forward to tackle these challenges.

In the realm of data consensus techniques, Latif \textit{et al.} \cite{latif2012realizing} developed the iRRR algorithm, which employs clustering and verification mechanisms to reject incorrect loop closures. Similarly, Sünderhauf \textit{et al.} \cite{sunderhauf2012switchable} propose a methodology that utilizes switchable constraints, thereby offering a dynamic system for scrutinizing and potentially rejecting specific loop closures, thus underlining the significance of data agreement within the graph.

\begin{figure*}[!ht]
    \includegraphics[width=\textwidth]{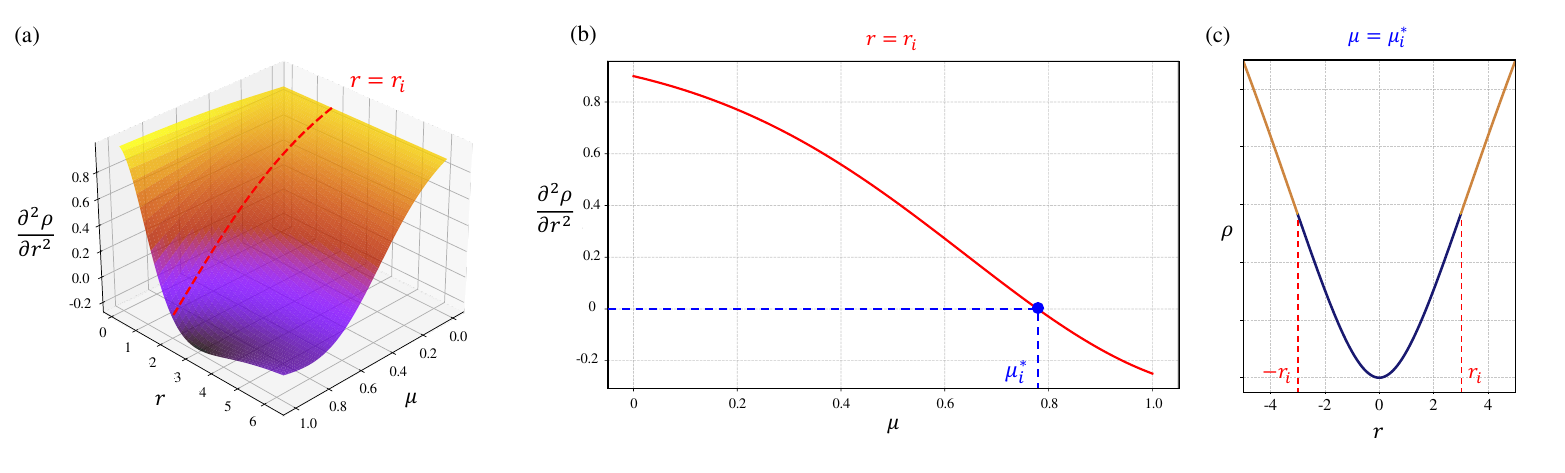}
    \caption{(a) The graph representing the second partial derivative of $\rho$ with respect to $r$. For the cross-section (red line) when $r=r_i$, we can obtain the ${\partial^2\rho}/{\partial r^2}$ graph for $\mu$ at a specific residual value. (b) The ${\partial^2\rho}/{\partial r^2}$ graph for $\mu$ when $r=r_i$. Points where the second partial derivative is 0 are identified as the points where the graph's convexity changes, and they are adopted as $\mu^*_i$. (c) The graph representing $\rho$ for the residual when $\mu=\mu^*_i$. The graph divides into concave (purple line) and convex (orange line) parts based on the $|r_i|$ value as a reference.}
    \label{fig:second_derivative}
\end{figure*}

In terms of mixture models designed for multi-modal instances, Olson \textit{et al.} \cite{olson2013inference} propose an alternative to traditional sum-mixture models. They advocate for the selection of maximum values from two probability density functions to achieve a similar distribution but with reduced computational load. Complementing this, Pfeifer \textit{et al.} \cite{pfeifer2021advancing} suggest a max-sum-mixture framework that, although computationally more demanding, benefits from the inclusion of non-linear terms, displaying enhanced precision in the results.

When focusing on robust kernel-based methodologies, M-estimator-based approaches have received considerable attention. 
M-estimator-based approaches introduce robust kernel $\rho$, forming the cost function $\sum _i \rho(r_i)$.
Various kernel function $\rho$ has been introduced, such as Huber \cite{Huber1992}, Cauchy, and Geman-McClure \cite{zhang1997parameter}.

Notably, Agamennoni \textit{et al.} \cite{agamennoni2015self} presents parameter tuning of M-estimators for handling data outliers, developing an automatic adjustment algorithm through the connection with elliptical probability distributions. Similarly, Chebrolu \textit{et al.} \cite{chebrolu2021adaptive} introduce an adaptive loss function \cite{Barron} for non-linear least squares problems, which features an automated kernel tuning based on the distribution of residuals. These researches expand the scope of existing robust kernels and improves accuracy without the need for manual parameter adjustments.

In addition, various approaches propose to enhance the robustness of SLAM. Lee \textit{et al}. \cite{lee2013robust} construct the pose graph SLAM using a Bayesian network and improved its performance using the expectation-maximization (EM) algorithm. Carlone and Calafiore \cite{carlone2018convex} introduce a convex relaxation-based pose graph optimization using the $l_1$ norm and the Huber loss function \cite{Huber1992}. Lajoie \textit{et al}. \cite{lajoie2019modeling} strengthen robustness against outliers by developing a DC-GM-based SLAM, drawing on a graphical model of pose graph and inlier-outlier selection, utilizing perceptual aliasing.

Shifting perspectives, Yang \textit{et al.} \cite{yang2020graduated} introduce GNC \cite{GNC} to specifically tackle outlier vulnerabilities in methods like semi-definite programming (SDP) and sums-of-squares (SOS) relaxations. Their method surpasses the performance of existing outlier-robust methods like RANSAC \cite{fischler1981random}. Building upon this, McGann \textit{et al.} \cite{mcgann2023robust} adapt GNC on online incremental SLAM \cite{kaess2011isam2} with their robust incremental Smoothing and Mapping (riSAM), which shows improvements over current outlier handling methodologies.

Regarding the topic of GNC optimization, conventional approaches for selecting control parameters have been mainly following predetermined update rules \cite{mcgann2023robust}, \cite{yang2020graduated}, \cite{le2021robust}, \cite{peng2023convergence}. 
This approach has been empirically proven to be effective but generally involves many steps and relies on heuristics. 

Building on that observation, Choi et al. \cite{choi2023adaptive} aimed to reduce the number of steps in the initial stages of GNC, particularly when non-convexity is low. They have examined the utility of shape functions as a more refined control mechanism. These shape functions offer a quicker yet well-disciplined GNC schedule during the early optimization phases. However, it is crucial to note that these methods remain inherently heuristic, implying room for further improvement due to the uncertainties and approximations intrinsic to heuristic approaches.

This paper proposes a novel approach to GNC-based methods, focusing on the intricate relationship between the robust kernel's control parameter and its non-convexity. In addition, we put forth a new policy for determining the steps of GNC, moving beyond the constraints of predetermined incremental changes.


\begin{figure*}[!ht]
    \includegraphics[width=\textwidth]{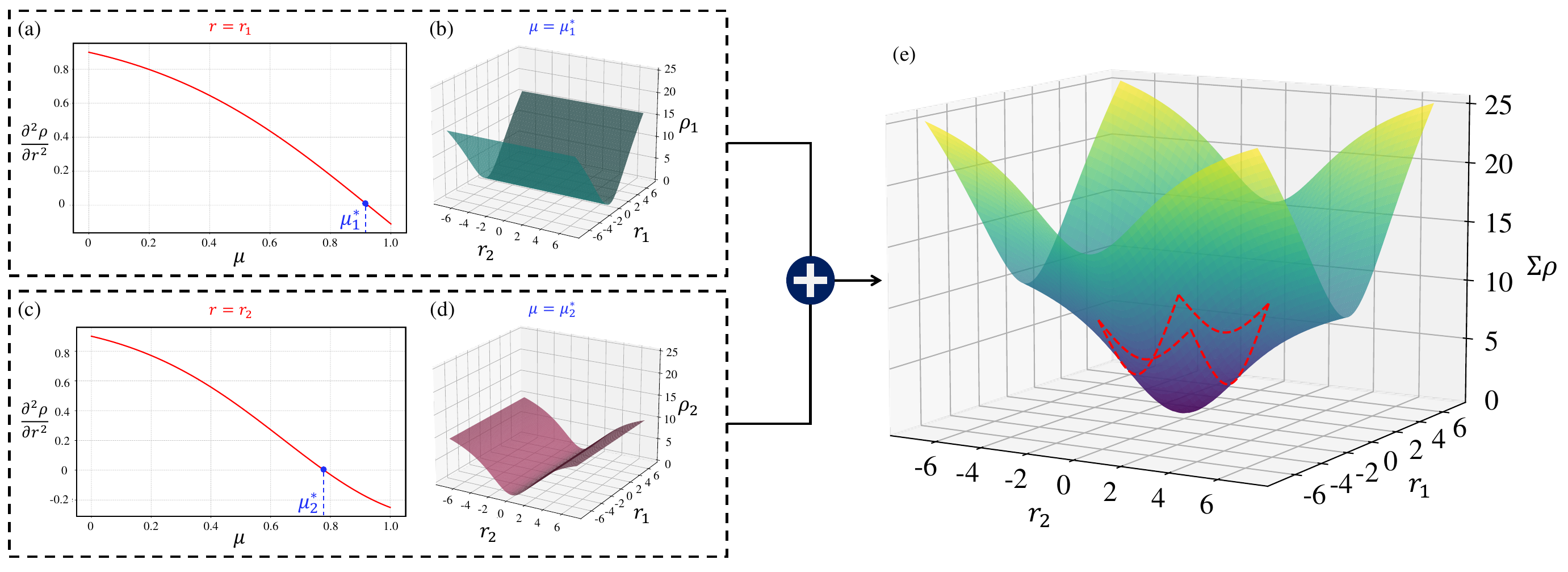}
    \caption{A simple example of convexity boundary for a pose graph with a factor node size of 2. Given nodes of the pose graph as \( \textbf{r} = \{r_1, r_2\} \), \( \mu^*_i \) can be calculated in the optimization step of \( r_i \). In this example, the initial guess is set to $r_1^0=2, r_2^0=3$. (a), (c) Visualization of \eqref{eq:SIG_second_derivative} when \( r = r_i \). At the convexity boundary point \( \mu^*_i \), this value becomes zero. (b), (d) Visualization of \eqref{eq:SIG_kernel} for each \( \mu^*_i \). \( \rho_i \) is only dependent on \( r_i \). (e) Visualization of the sum of all \( \rho_i \), denoted \( \Sigma \rho \). The \( \Sigma \rho \) value for \( \pm r_i \) is convex within its boundary (red dotted line). In this example, $\Sigma \rho$ visualized using 3D plot with axes $r_1, r_2$ and the convexity boundary is $-2<r_1<2$, and $-3<r_2<3$.
}
    \label{fig:convex_region}
\end{figure*}

\section{Methodology}

In this section, we present the details of our novel methodology, which seeks to optimize the control parameter $\mu$ by leveraging the convex properties of the Scale Invariant Graduated (SIG) kernel in GNC. We start by clarifying the key principles behind the GNC method and the crucial role of the control parameter $\mu$ and \textit{GNC schedule}. We then explore the unique attributes of the SIG kernel, explaining how its inherent convexity can contribute to enhancing the robustness and efficiency of non-convex optimization algorithms. Building on these characteristics, we then propose our novel GNC schedule to harness these benefits.

\subsection{Control Parameter $\mu$ and GNC Schedule}
The GNC method serves as a valuable methodology for optimizing non-convex functions. Originating from an initial convex state, this approach incrementally elevates the non-convexity of the function intending to mitigate sensitivities to initial conditions and reduce the likelihood of convergence to local minima. Those kernels use the control parameter $\mu$ to vary the shape of the kernel. GNC methods incrementally adjust the $\mu$ value to increase non-convexity. The policy of changing the $\mu$ value is called a \textit{GNC schedule}. When employed in frameworks such as PGO, robust kernels, such as truncated least squares (TLS) \cite{yang2020graduated}, are often utilized to diminish the influence of outliers.

It is important to note that the choice of the GNC schedule could substantially impact the efficiency and stability of the method. Should the GNC schedule adjust in excessively fine increments, computational costs may inflate. Conversely, a schedule that changes the $\mu$ value too rapidly carries the risk of trapping the optimization in local minima. The traditional approach, which relies on empirically derived schedules, can fail in certain scenarios since it has not always embraced the most efficient strategies in terms of computational efficiency and robustness. Therefore, the careful selection and adjustment of the GNC schedule is a critical consideration in optimizing the performance of the method. The efficient GNC schedule has the potential to enhance the robustness and efficiency of non-convex optimization algorithms. To harness this potential, it would be beneficial to explore suggestions and insights for GNC schedule adjustments that promote both computational efficiency and algorithmic reliability.

The difference between the traditional method and our proposed method can be briefly seen in Fig. \ref{fig:step_compare}. As observed in the upper image of Fig. \ref{fig:step_compare}, in noisy datasets, the traditional method can fail because it doesn't schedule $\mu$ values close to \textit{convexity boundary points} $\mu^*$. $\mu^*$ distinguishes between the $\mu$ value where the kernel function is globally convex and those where it isn't. In the lower image of Fig. \ref{fig:step_compare} our proposed method schedules only the $\mu$ values closest to the original function while maintaining the convex property of the kernel function. This demonstrates greater robustness and computational efficiency compared to traditional methods.

\subsection{Convex Property of SIG Kernel}
Traditional robust kernels, such as the Geman-McClure (GM) kernel, present challenges due to their inconsistent convexity properties across the entire domain. To address this limitation, McGann \textit{et al.} \cite{mcgann2023robust} introduced the Scale Invariant Graduated (SIG) kernel, which maintains convexity across its domain. We employ the SIG kernel to take advantage of its scale-invariant properties. 

The SIG kernel is defined as:
\begin{equation}
    \rho(r;\mu) = \frac{1}{2} \frac{c^2 r^2}{c^2 + (r^2)^\mu}
    \label{eq:SIG_kernel}
\end{equation}
When \(\mu=0\), this equation manifests as a convex quadratic function, and as the value of \(\mu\) increases, it approximates the behavior of the Geman-McClure kernel more closely, eventually becoming identical when $\mu=1$.

In a given context with an initial value $r_i$, we aim to identify the range where the SIG kernel's convexity is preserved and optimize within this range. This can be determined by taking the second derivative of the SIG kernel with respect to $r$ and identifying points where the value of the second derivative becomes zero. The expression for the second derivative is as follows:
\begin{multline}
    \quad\quad\quad \frac{ \partial^2 \rho }{ \partial r^2} = \frac{4c^2 \mu^2 (r^2)^{2\mu}}{(c^2+(r^2)^\mu)^3} - \frac{2c^2 \mu^2 (r^2)^\mu}{(c^2+(r^2)^\mu)^2} \\
    - \frac{3c^2 \mu (r^2)^\mu}{(c^2+(r^2)^\mu)^2} + \frac{c^2}{c^2+(r^2)^\mu} \quad\quad\quad
    \label{eq:SIG_second_derivative}
\end{multline}

This is visually represented in Fig. \ref{fig:second_derivative}-(a). Plotting $\mu$ against function values for a given $r_i$ results in a curve resembling Fig. \ref{fig:second_derivative}-(b). This allows us to identify a specific $\mu^*_i$ where the value of the second derivative becomes zero.

Upon fixing $\mu$ at $\mu^*_i$, the SIG kernel exhibits the following characteristics with respect to the range of $r$, as clearly illustrated in Fig. \ref{fig:second_derivative}-(c):
\begin{itemize}
    \item \(0 \leq r < r_0\): The kernel function remains convex.
    \item \(r = r_0\): An inflection point occurs.
    \item \(r > r_0\): The kernel function becomes concave.
\end{itemize}

Likewise, the kernel's behavior varies depending on $\mu$, as evident from Fig. \ref{fig:second_derivative}-(a) and Fig. \ref{fig:second_derivative}-(b):

\begin{itemize}
    \item \(0 \leq \mu < \mu^*\): The function can remain convex even for $r$ greater than $r_i$.
    \item \(\mu = \mu^*\): For each \(r_0\), the function is convex only within a range smaller than $r_i$.
    \item \(\mu^* < \mu \leq 1\): Convexity is not guaranteed even for ranges smaller than $r_i$.
\end{itemize}

Through this approach, we are able to harness the convexity boundaries of the SIG kernel to furnish a more robust and efficient solution to non-convex optimization problems. This is anticipated to alleviate dependency on initial conditions and mitigate issues related to local minima. Although the initial findings are promising, further investigations are warranted to corroborate these observations and optimize the \(\mu^*\) selection.

\subsection{Efficient GNC Schedule}


\begin{algorithm}[t]
\SetAlgoNlRelativeSize{0}
\caption{Efficient GNC schedule}
\label{alg:schedule}
\KwIn{Factors \(\mathscr{F}\) }
\KwOut{Optimized Factors \(\mathscr{F}_{optimized}\) }
\BlankLine

all \( \mu_i = 0 \) \\
all \( \text{step}_i = 0 \) \\

\While{not Converged(all \( \mu_i \), 1)}{
  \ForEach{factor in \(\mathscr{F}\)}{
    \If{\text{init\_step[factor\_id]} == 2}{
      step \( \leftarrow \) init\_step
    }
    compute \( r_i \) \\
    \uIf{step == 0}{
      \( \mu^*_i = 0 \)
    }
    \uElseIf{step == 1 and \( r_i < \chi^n(0.9) \)}{
      Compute $\mu_i$ such that $\text{second\_derivative}(r_i, \mu_i)=0$ \\
    \( \mu^*_i \leftarrow \min(\mu_i, 1) \)
    }
    \Else{
      \( \mu^*_i = 1 \)
    }
    step \( += 1 \)
  }
  \(\mathscr{F}_{optimized}\) \( \leftarrow \) OptimizeAllFactors(\(\mathscr{F})\) \\
  
  \ForEach{factor in \(\mathscr{F}\)}{
    compute \( r_i \) \\
    \If{\( r_i > \chi^n(0.9) \)}{
      init\_step[factor\_id] \( = 2 \)
    }
  }
}
\end{algorithm}

In this study, we strategically leverage the convexity properties of the Scale Invariant Graduated (SIG) kernel to make informed choices for the control parameter \(\mu\). This approach has proven effective in maintaining the kernel's convexity over a specified range of residuals.

The simple example of kernel function for a pose graph with a node size of 2 is represented in Fig. \ref{fig:convex_region}. As demonstrated in Fig. \ref{fig:convex_region}-(a) and \ref{fig:convex_region}-(c), we can identify a \( \mu^* \) for each residual that ensures the convexity of the SIG kernel. Across the entire node $\textbf{r}$'s parameter space, the kernel function $\rho$ is visualized as shown in \ref{fig:convex_region}-(b) and \ref{fig:convex_region}-(d). Given that the sum of convex functions is also convex, as mathematically proven \cite{boyd2004convex}, we can further establish a \textit{convexity boundary} for the sum of the kernel functions, as visualized in Fig. \ref{fig:convex_region}-(e).

In essence, for each residual, we can guarantee a boundary where \( r \) values are smaller than the initial \( r_i \) values, and within which convexity is maintained. By optimizing within the convex boundary, the optimal solution is assured to be a global optimum \cite{boyd2004convex}, as is well-understood in convex optimization. This also implies that for \( \mu \) values smaller than \( \mu^* \), where the kernel function maintains its convexity, explicit optimization is not always necessary. Moreover, convex optimization generally converges faster than its non-convex counterparts, offering computational speed advantages. The selected \( \mu \) value, being in the range where convexity is maintained, minimizes the risk of falling into local minima. For larger \( \mu \) values beyond \( \mu^* \), the kernel function turns concave, and we cannot guarantee the convexity of the objective function. Taking these factors into account, we conduct GNC optimization steps at the boundary \( \mu^* \), thereby enhancing computational efficiency while preserving robustness.

Additionally, for instances where the residual of a factor is considerably large—exceeding the 0.9 in a chi-square distribution—we categorize it as an \textit{strong outlier}. In these cases, we bypass the calculated value of \(\mu^*\) and directly set \(\mu\) to 1. Should the residual remain large even after iterative optimization, we treat it as an outlier in subsequent optimization stages and initialize \(\mu\) to 1. This strategy aims to mitigate the influence of identified outliers on the optimization process.
We summarize our GNC schedule in Alg. \ref{alg:schedule}.


\begin{figure}[!t]
    \includegraphics[width=0.486\textwidth]{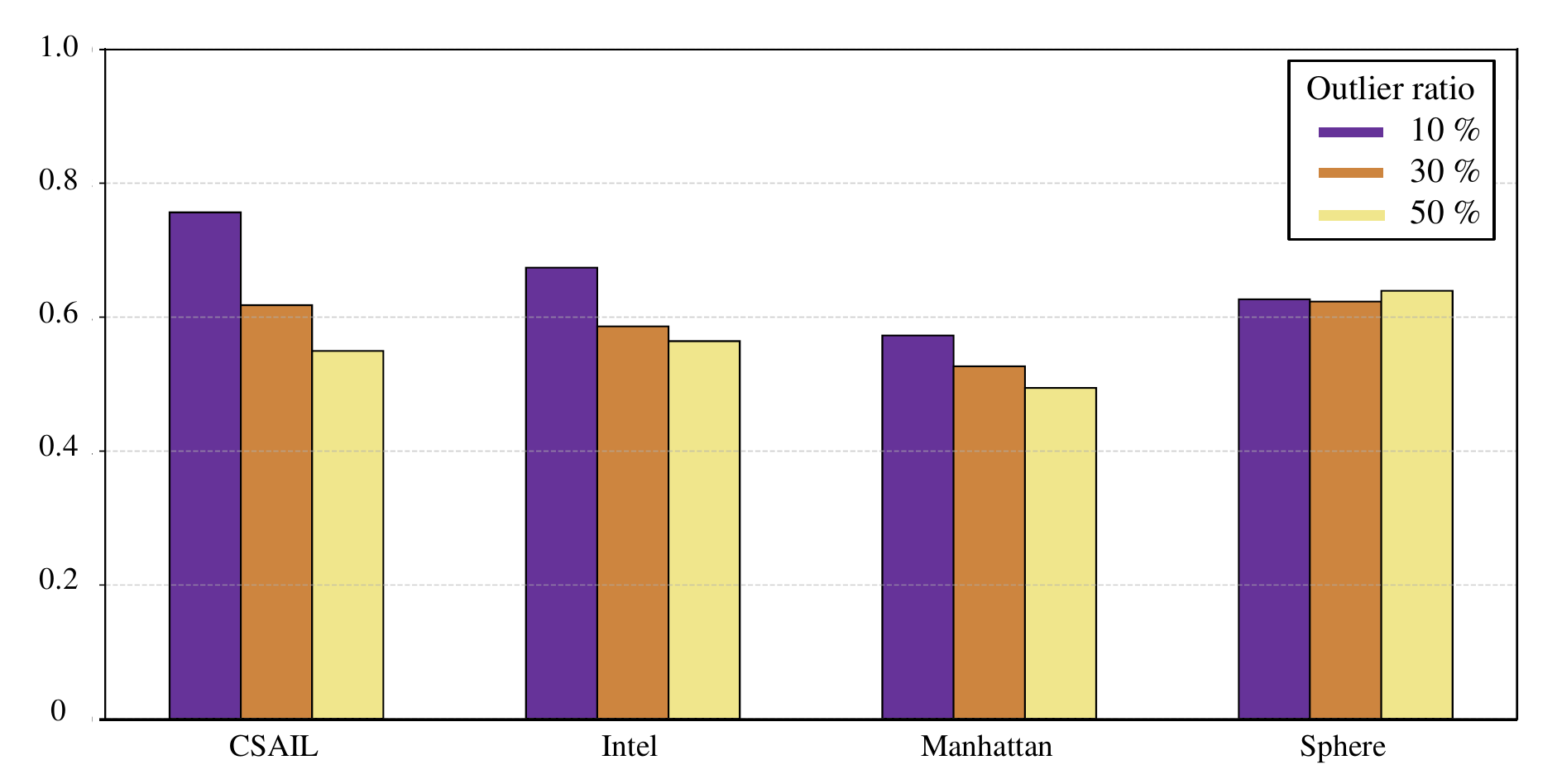}
    \caption{The bar graph illustrating the comparative runtime of our proposed approach versus riSAM when the datasets experience outlier perturbations at rates of 10\%, 30\%, and 50\%.}
    \label{fig:Results_graph}
\end{figure}

\begin{table}[]
\centering
\caption{Results of the first experiment comparing precision and recall between our method and riSAM across various datasets. Performance is similar in all but the Sphere2500 dataset, where our method exhibits superior performance, highlighted in bold.}
\label{tab:table1}
\begin{tabular}{|cc|cc|cc|}
\hline
\multicolumn{2}{|c|}{Metric}                                                                                   & \multicolumn{2}{c|}{Precision}       & \multicolumn{2}{c|}{Recall}          \\ \hline
\multicolumn{1}{|c|}{Dataset}                    & \begin{tabular}[c]{@{}c@{}}Outlier\\ Ratio [\%]\end{tabular} & \multicolumn{1}{c|}{riSAM}  & Ours   & \multicolumn{1}{c|}{riSAM}  & Ours   \\ \hline
\multicolumn{1}{|c|}{\multirow{3}{*}{CSAIL}}     & 10                                                          & \multicolumn{1}{c|}{1.0}      & 1.0      & \multicolumn{1}{c|}{0.9922} & 0.9922 \\ \cline{3-6} 
\multicolumn{1}{|c|}{}                           & 30                                                          & \multicolumn{1}{c|}{1.0}      & 1.0      & \multicolumn{1}{c|}{0.9922} & 0.9922 \\ \cline{3-6} 
\multicolumn{1}{|c|}{}                           & 50                                                          & \multicolumn{1}{c|}{1.0}      & 1.0      & \multicolumn{1}{c|}{0.9922} & 0.9922 \\ \hline
\multicolumn{1}{|c|}{\multirow{3}{*}{Intel}}     & 10                                                          & \multicolumn{1}{c|}{1.0}      & 1.0      & \multicolumn{1}{c|}{1.0}      & 1.0      \\ \cline{3-6} 
\multicolumn{1}{|c|}{}                           & 30                                                          & \multicolumn{1}{c|}{1.0}      & 1.0      & \multicolumn{1}{c|}{1.0}      & 1.0      \\ \cline{3-6} 
\multicolumn{1}{|c|}{}                           & 50                                                          & \multicolumn{1}{c|}{1.0}      & 1.0      & \multicolumn{1}{c|}{1.0}      & 1.0      \\ \hline
\multicolumn{1}{|c|}{\multirow{3}{*}{Manhattan}} & 10                                                          & \multicolumn{1}{c|}{1.0}      & 1.0      & \multicolumn{1}{c|}{1.0}      & 1.0      \\ \cline{3-6} 
\multicolumn{1}{|c|}{}                           & 30                                                          & \multicolumn{1}{c|}{1.0}      & 1.0      & \multicolumn{1}{c|}{1.0}      & 1.0      \\ \cline{3-6} 
\multicolumn{1}{|c|}{}                           & 50                                                          & \multicolumn{1}{c|}{0.9995} & 0.9995 & \multicolumn{1}{c|}{1.0}      & 1.0      \\ \hline
\multicolumn{1}{|c|}{\multirow{3}{*}{Sphere}}    & 10                                                          & \multicolumn{1}{c|}{1.0}      & 1.0      & \multicolumn{1}{c|}{\textbf{0.9984}} & 0.9980 \\ \cline{3-6} 
\multicolumn{1}{|c|}{}                           & 30                                                          & \multicolumn{1}{c|}{1.0}      & 1.0      & \multicolumn{1}{c|}{0.9967} & \textbf{0.9976} \\ \cline{3-6} 
\multicolumn{1}{|c|}{}                           & 50                                                          & \multicolumn{1}{c|}{0.9992} & \textbf{1.0}     & \multicolumn{1}{c|}{0.9853} & \textbf{0.9967} \\ \hline
\end{tabular}
\end{table}

\section{Experiments}
In this section, we present the evaluation procedures and results of our proposed method using the pose graph optimization datasets. The experiments are divided into two categories. The first experiment is designed to test the method's capability in handling false loop closures due to incorrect spatial recognition, using the CSAIL \cite{csaildata}, Intel\cite{intelmanhattandataset}, Manhattan3500\cite{manhattandataset}, \cite{intelmanhattandataset}, and Sphere2500\cite{spheredataset} datasets. The second experiment utilizes the ground truth of the Manhattan dataset to create a more challenging dataset, taking sensor errors into account and considering scenarios with noisy relative pose measurements.

\subsection{First Experiment}
The initial experiment artificially introduces outliers or false loop closures in the dataset, emulating scenarios where far-apart poses are incorrectly recognized as the same location. We simulate this by adding incorrect false loop closures at a given outlier ratio. The runtime and capability to detect false outliers of both riSAM and our method are compared using precision and recall metrics.

\subsubsection{Time Comparison} Firstly, we compare the execution time of the methods. Fig. \ref{fig:Results_graph} compares the average runtime of riSAM and our method, each run five times on the same data. Based on the average time of riSAM, set as a benchmark of 1.0, we calculate the time ratio of our method. Our method requires approximately 60\% of riSAM's execution time, performing increasingly better as the dataset size increased.


\begin{figure}[!t]
    \includegraphics[width=0.486\textwidth]{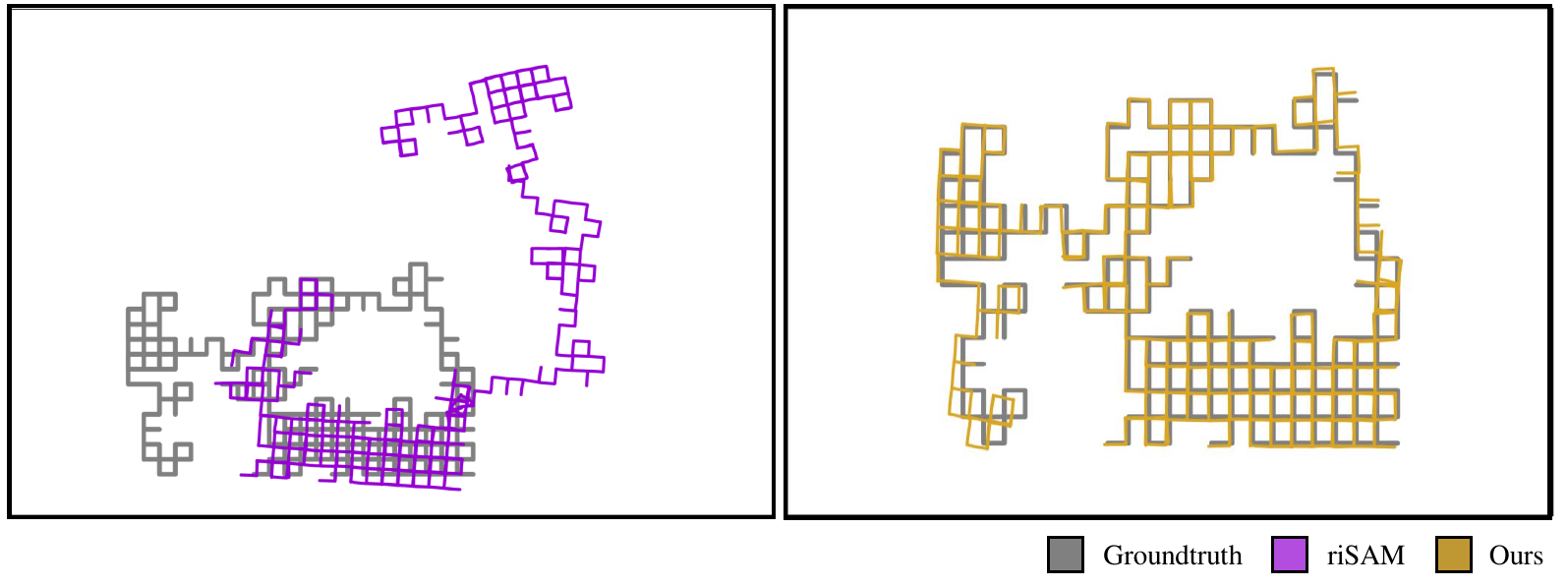}
    \caption{Results of experimental on \textit{challenging} case of Manhattan3500 dataset with introduced noise. Both images contain ground truth trajectory, marked with a gray line. The left image represents the result of riSAM, and the right image represents the result of our method.}
    \label{fig:Results_trajectory}
\end{figure}

\subsubsection{Detecting False Outliers} Next, we evaluate the ability to detect false outlier loops. Outliers and inliers are classified based on the error measured after optimization, compared against a chi-squared distribution threshold of 0.95. This is consistent with the minimum error value used for defining outliers during the dataset generation \cite{mcgann2023robust}. As illustrated in Table \ref{tab:table1}, our method shows robustness comparable to riSAM. Particularly noteworthy is its performance on the Sphere2500 dataset, which contains 50\% outliers. While riSAM's trajectory tends to crash under these conditions, our method effectively filters out false loop closures, maintaining a coherent and correct trajectory. This marked difference is visible in Fig. \ref{fig:sphere_results}.

\subsection{Second Experiment}
The second experiment employs a more challenging dataset, adding Gaussian noise to the odometry and false loop relative poses in the ground truth of the Manhattan dataset.

\subsubsection{Data Generation} Random Gaussian noise is added to all odometry measurements in the Manhattan3500 ground truth dataset. Subsequently, a set ratio of existing loops is randomly chosen to have added noise. In our case, 30\% of the loops are selected, creating a dataset that challenges both riSAM and our method in filtering out false loops effectively.

\subsubsection{Comparison} We compare the results of running riSAM and our method on the generated datasets. Success is defined as cases where both precision and recall are 1.0. As shown in Table \ref{tab:table 2}, our method succeeds in more datasets than riSAM. This is because our method avoids local minimum traps, restoring correct trajectories, while riSAM sometimes fails to do so. Additionally, as seen in Table \ref{tab:table 3}, in certain cases, our method not only reduces execution time but also generally shows favorable performance over riSAM in key metrics such as RPE, ATE, precision, and recall. In the specific scenarios highlighted in Table \ref{tab:table 3}, our method shows favorable performance over riSAM in key metrics. As evidenced in Fig. \ref{fig:Results_trajectory}, while riSAM's trajectory tends to crash under these conditions, our method maintains a coherent and accurate path. It's worth noting that, as indicated in Table \ref{tab:table 2}, there are multiple such cases where our method outperforms riSAM.

\section{Discussion and Future Work}

The results presented in the previous section demonstrate that applying the GNC method to PGO significantly improves computational efficiency while either maintaining or enhancing robustness. In comparison to riSAM, our approach displays substantially faster computation speeds. This advantage becomes increasingly pronounced, especially as challenging false loops are introduced, thereby indicating greater robustness. By leveraging the properties of convex functions, our method minimizes the number of iteration steps and optimizes solutions in a way that increases robustness even in the presence of outliers.

\begin{table}[!t]
\caption{Results of the second experiment conducted with 50 generated datasets. Our method succeeds in a greater number of cases compared to riSAM.}
\centering
\label{tab:table 2}
\begin{tabular}{|c|c|c|c|}
\hline
              & Success                          & Fail        & Success Rate [\%] \\ \hline
riSAM         & 15                               & 35          & 30                \\ \hline
\textbf{Ours} & \textbf{19}                      & \textbf{31} & \textbf{38}       \\ \hline
\end{tabular}
\end{table}

\begin{table}[]
\caption{Metrics measured for cases where riSAM experiences a crash. Our method not only reduces execution time but also outperforms riSAM across all evaluated metrics, including precision, recall, RPE, and ATE.}
\centering
\label{tab:table 3}
\begin{tabular}{|c|c|c|c|c|c|}
\hline
Metric        & Precision    & Recall       & RPE {[}\%{]}   & ATE {[}m{]}    & Runtime {[}s{]} \\ \hline
riSAM         & 0.9921       & 0.9672       & 45.536         & 23.636         & 58.28           \\ \hline
\textbf{Ours} & \textbf{1.0} & \textbf{1.0} & \textbf{3.217} & \textbf{1.446} & \textbf{43.05}  \\ \hline
\end{tabular}
\end{table}

Looking ahead, we identify two primary avenues for future work. Firstly, further refinement of the algorithm is necessary. Although our proposed method reduces the likelihood of falling into local minima by optimizing at points where convexity is maintained, the potential still exists for encountering local minima in non-convex settings. Identifying the regions where local minima may form as we increase the value of $\mu$ could allow us to preemptively optimize and avoid these pitfalls. Secondly, while our tests have been on simulation datasets, applying our method to real-world data is essential for further validation.


\newpage

{\small
\bibliographystyle{IEEEtran}
\bibliography{bib}
}

\end{document}